\documentclass{article}

\usepackage{amssymb}
\usepackage{centernot}
\usepackage{pifont}
\usepackage{tcolorbox}
\usepackage{rotating}
\tcbuselibrary{breakable}
\usepackage[preprint]{neurips_2026}

\usepackage[utf8]{inputenc} 
\usepackage[T1]{fontenc}    
\usepackage{hyperref}       
\usepackage{url}            
\usepackage{booktabs}       
\usepackage{amsfonts}       
\usepackage{amssymb}        
\usepackage{amsmath}        
\usepackage{nicefrac}       
\usepackage{microtype}      
\usepackage{xcolor}         
\usepackage{graphicx}
\usepackage{subcaption}
\usepackage{svg}
\usepackage{algorithm,algorithmic}
\usepackage{amsthm}

\usepackage{multirow}
\usepackage{tikz}
\usepackage{xcolor}
\usepackage{calc}
\usepackage{fancyvrb}
\usepackage{pifont}
\usepackage[dvipsnames]{xcolor}

\definecolor{HeroViolet}{HTML}{746ACD}
\definecolor{HeroOrchid}{HTML}{BE8EEA}
\definecolor{HeroPink}{HTML}{E28FD3}
\definecolor{HeroGold}{HTML}{F0A81E}

\usepackage{expl3}
\ExplSyntaxOn
\int_new:N \g_hero_total_int  \int_new:N \g_hero_pos_int
\cs_new_protected:Npn \__hero_space: { \space }
\cs_new_protected:Npn \__hero_setcol:
 { \fp_set:Nn \l_tmpa_fp
    { \int_compare_p:nNn {\g_hero_total_int} > {1}
       ? { 100*(\g_hero_pos_int-1)/(\g_hero_total_int-1) } : { 50 } }
   \int_set:Nn \l_tmpa_int { \fp_to_int:n {\l_tmpa_fp} }
   \int_compare:nNnTF {\l_tmpa_int} < {50}
    { \colorlet{heropos}{HeroViolet!\int_eval:n{100-2*\l_tmpa_int}!HeroOrchid} }
    { \colorlet{heropos}{HeroOrchid!\int_eval:n{100-2*(\l_tmpa_int-50)}!HeroPink} } }
\NewDocumentCommand{\sweep}{m}   
 { \tl_set:Nn \l_tmpa_tl {#1} \tl_replace_all:Nnn \l_tmpa_tl { ~ } { \__hero_space: }
   \tl_map_inline:Nn \l_tmpa_tl
    { \token_if_eq_meaning:NNTF ##1 \__hero_space: { \__hero_space: }
        { \int_gincr:N \g_hero_pos_int \__hero_setcol: \textcolor{heropos}{##1} } } }
\NewDocumentCommand{\gold}{m}    
 { \tl_set:Nn \l_tmpa_tl {#1} \tl_replace_all:Nnn \l_tmpa_tl { ~ } { \__hero_space: }
   \tl_map_inline:Nn \l_tmpa_tl
    { \token_if_eq_meaning:NNTF ##1 \__hero_space: { \__hero_space: }
        { \int_gincr:N \g_hero_pos_int \textcolor{HeroGold}{##1} } } }
\NewDocumentCommand{\herobegin}{m}
 { \int_gset:Nn \g_hero_total_int {#1} \int_gzero:N \g_hero_pos_int }
\ExplSyntaxOff

\definecolor{wandbgold}{HTML}{FFBE00}

\theoremstyle{definition}

\theoremstyle{remark}

\definecolor{creamfill}{RGB}{255,249,235}   
\definecolor{creamstroke}{RGB}{210,180,140} 

\title{\herobegin{23}\sweep{Going }\gold{Deep}\sweep{er With }\gold{Mem}\sweep{ories}}

\author{%
  Afshin Khadangi\\
  SnT, University of Luxembourg\\
  \texttt{afshin.khadanki@uni.lu}\\
  \href{https://afshin.xyz}{\texttt{afshin.xyz}}
}

\begin{document}

\maketitle


\begin{minipage}{\linewidth}
    \centering
    \includegraphics[
        width=\linewidth,
        keepaspectratio
    ]{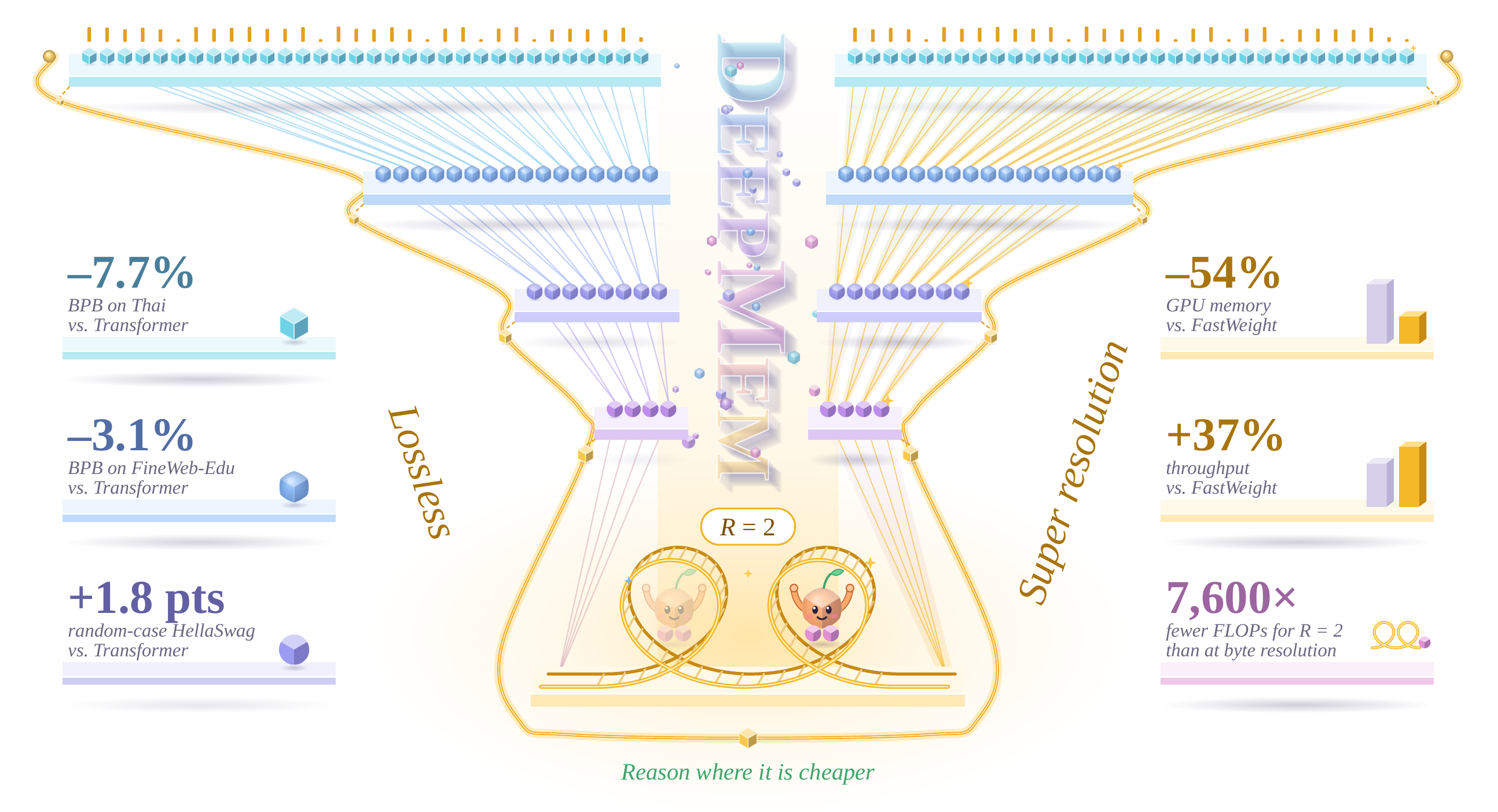}

    \captionof{figure}{
        \textsc{DeepMem}. Bytes descend into fewer memories while a lossless carrier (gold) keeps every byte exact. At the 16-position bottleneck, one shared state is refined twice ($R{=}2$) at ${\sim}7{,}600\times$ fewer FLOPs than at full byte resolution, and super-resolution restores byte-level states. Side panels: gains vs.\ a matched Transformer (left) and the FastWeight baseline (right).
    }
    \label{fig:ca_overview}
\end{minipage}

\vspace{0.8em}

\begin{abstract}
Byte level language models avoid tokenizer dependence and provide universal coverage, but processing every byte at full resolution is expensive. We introduce \textbf{\textsc{DeepMem}}, a causal multiscale architecture that separates semantic compression from exact byte preservation. A learned invertible subband transform constructs a hierarchy of shorter sequences, while a lossless carrier and explicit detail sidebands preserve byte identity. Fast weight memory blocks operate along the encoder, with iterative shared state computation at the compressed bottleneck, while an attention based super resolution decoder restores byte resolution representations. \textsc{DeepMem} achieves stronger language modeling performance than a matched byte level Transformer on FineWeb-Edu and PG19, and obtains the best macro average performance across the evaluated FLORES languages. It also improves PIQA accuracy while remaining competitive on code modeling, multiple choice reasoning, and byte level corruption robustness. Bottleneck refinement requires approximately $7{,}600\times$ fewer counted FLOPs than equivalent full resolution reasoning, while maintaining competitive evaluation throughput.
\end{abstract}

\section{Introduction}

Modern language models typically process sequences produced by fixed subword tokenizers. Byte-pair encoding and unigram tokenization shorten sequences and make Transformer training practical \citep{sennrich2016neural,kudo2018sentencepiece,vaswani2017attention}, but require choosing a vocabulary and segmentation in advance. Tokenizers can fragment rare words, names, code, and morphologically rich forms; compress languages and scripts unevenly; and react discontinuously to small character perturbations \citep{rust2021tokenizer,xue2022byt5}. They therefore introduce an external representation bottleneck despite bytes already providing a universal digital representation.

Byte-level models remove this dependency. A fixed vocabulary of 256 symbols represents arbitrary Unicode text and other digital data without unknown tokens while exposing spelling, morphology, punctuation, and formatting. ByT5 demonstrated strong byte-level representations and robustness to character noise \citep{xue2022byt5}, while CANINE and Charformer improved efficiency by shortening character or byte sequences before deeper processing \citep{clark2022canine,tay2022charformer}. The remaining challenge is sequence length: documents contain many more bytes than subword tokens, making full-resolution computation expensive.


Recent work addresses this cost through internal compression. MEGABYTE separates global patch computation from local byte prediction \citep{yu2023megabyte}; SpaceByte creates larger blocks near likely word boundaries \citep{slagle2024spacebyte}; and MambaByte uses a selective state-space backbone \citep{wang2024mambabyte}. Byte Latent Transformer forms entropy-based variable-length patches \citep{pagnoni2025blt}, MrT5 dynamically removes redundant positions \citep{kallini2025mrt5}, and Multiscale Byte Language Models use multiple autoregressive stages \citep{egli2025mblm}. Other hierarchical models combine byte- and word-level computation \citep{neitemeier2025hierarchical}, while recent work explores diffusion and speculative mechanisms for faster byte generation \citep{kallini2026fastblt}. These approaches show that tokenizer-free modeling is viable, but its efficiency depends strongly on how computation is distributed across resolutions.

Hierarchical computation also appears beyond byte modeling. Funnel Transformers shorten hidden sequences and reinvest the savings in deeper representations \citep{dai2020funnel}, while Hourglass Transformers combine autoregressive downsampling and upsampling \citep{nawrot2022hourglass}. MEGABYTE uses coarse representations to coordinate local decoders, and Autoregressive U-Nets learn hierarchies from bytes toward abstract units \citep{yu2023megabyte,videau2025aunet}. All avoid performing every layer at full resolution, but pooling, deletion, merging, and patching create a tension between efficiency and fidelity: compressed states must preserve exact byte information, rely on a separate local path, or reconstruct detail later.

We address this tension using multirate signal-processing principles. Laplacian pyramids represent signals through coarse approximations and residual bands \citep{burt1983laplacian}. The lifting scheme implements invertible wavelet transforms through learned predict and update operations while preserving an exact algebraic inverse \citep{sweldens1998lifting}. U-Net similarly combines a contracting path with an expanding path and skip connections for fine-resolution recovery \citep{ronneberger2015unet}. These ideas suggest that semantic computation can be compressed aggressively when exact byte information is preserved separately rather than forced through a narrow bottleneck.

Applying subband analysis to autoregressive modeling requires explicit causal constraints. A feature summarizing a span $[a,b]$ may influence only positions whose available context includes $b$. Likewise, a parent derived from two children cannot be copied to the earlier child without leaking information from the later one. Causality must therefore govern the analysis transform, synthesis path, and skip connections—not only individual attention or recurrent layers.

We introduce  \textbf{\textsc{DeepMem}} (Fig. \ref{fig:ca_overview}), a causal lossless subband architecture for byte-level autoregressive language modeling. It maintains a learned semantic stream and a lossless carrier stream. The semantic stream is progressively shortened to capture predictive structure, while each byte is mapped to a fixed eight-dimensional binary carrier and recursively transformed through learned lifting steps. Each level produces a coarse carrier and an explicit detail sideband whose hierarchy reconstructs the original carrier exactly in real arithmetic. Semantic compression therefore need not discard or implicitly memorize byte identity.

The encoder performs memory based mixing at progressively shorter resolutions. Fast-weight and linear attention models interpret sequence processing as recurrent associative memory, where a controller writes key--value associations into a rapidly changing state \citep{schmidhuber1992fastweights,katharopoulos2020transformers,schlag2021linear}. Gated linear attention adds data-dependent forgetting and chunk-parallel training \citep{yang2024gla}. We use this causal memory along the encoding path and in a two step shared state refinement module at the final semantic bottleneck. An attention based super-resolution decoder then restores byte resolution representations from delayed parent proposals, encoder skips, and exact carrier features.

This separation gives each path a distinct role. The semantic encoder performs deep computation at reduced lengths while retaining a sufficiently wide coarse representation for linguistic abstraction. The carrier preserves fine-grained byte information explicitly. The decoder combines compressed semantic
representations with locally available carrier
features and encoder skips to construct
high-resolution states for next-byte prediction. The method thus differs from single segmentation patch models and lossy hourglass architectures whose compressed states must retain all relevant local detail.

At approximately 100 million parameters, the model achieves $1.0134$ BPB on FineWeb-Edu, versus $1.0457$ for a matched raw-byte Transformer and $1.0187$ for a FastWeight baseline. It also improves on PG19 ($1.4402$ versus $1.4635$ BPB) and obtains the best macro-average across eight FLORES languages ($1.9694$ versus $2.0082$ BPB), with a modest tradeoff on Stack-smol code. It remains competitive on HellaSwag, PIQA, and ARC and is robust to repeated-character, Antspeak, uppercase, and random-case perturbations.

\paragraph{Contributions.}
Our main contributions are:
\begin{itemize}
\item a learned, recursively invertible subband decomposition separating a compressed semantic hierarchy from an exact byte carrier and explicit detail sidebands;
\item causal analysis, delayed super resolution synthesis, and skip-fusion rules that prevent completed span features from leaking to earlier positions;
\item a hybrid fast-weight memory encoder with two step shared state bottleneck refinement and an attention based super resolution decoder; and
\item improved general domain, long form, and multilingual BPB over a matched raw byte Transformer.
\end{itemize}

\section{Method}
\label{sec:method}

\subsection{Overview and Notation}
\label{sec:method-overview}

Let $\mathbf{x}=(x_0,\ldots,x_{T-1})$ be a sequence of raw bytes, where
$x_t\in\{0,\ldots,255\}$. Our model constructs a dyadic hierarchy with levels
$\ell=0,\ldots,L$, where level $0$ is the original byte resolution and each
successive level halves the sequence length. The architecture maintains two
parallel representations:

\begin{enumerate}
\item a learned semantic stream $\mathbf{S}^{(\ell)}$, which performs
predictive contextual computation; and
\item a lossless carrier stream $\mathbf{K}^{(\ell)}$, which preserves exact
byte identity through an invertible subband transform.
\end{enumerate}

At the byte level, the semantic representation is obtained from a learned
embedding table $\mathbf{E}$, while the carrier is a fixed bipolar
representation of the eight bits of each byte:
\begin{equation}
\label{eq:input-streams}
\begin{aligned}
\mathbf{S}^{(0)}_t &= \mathbf{E}[x_t], \quad
\mathbf{K}^{(0)}_{t,r}
= 2\mathbf{1}\!\left[
\left(
\left\lfloor x_t/2^r\right\rfloor \bmod 2
\right)=1
\right]-1,\\
&\qquad \qquad \qquad r\in\{0,\ldots,7\}.
\end{aligned}
\end{equation}
Thus, $\mathbf{K}^{(0)}_t\in\{-1,+1\}^{8}$ is a parameter-free injective
encoding of $x_t$.

At level $\ell$, the sequence length, semantic width, and carrier width are
denoted by $T_\ell$, $d_\ell$, and $c_\ell$, respectively:
\begin{equation}
\label{eq:pyramid-shapes}
T_\ell=\frac{T}{2^\ell},
\quad
\mathbf{S}^{(\ell)}\in\mathbb{R}^{T_\ell\times d_\ell},
\quad
\mathbf{K}^{(\ell)}\in\mathbb{R}^{T_\ell\times c_\ell},
\quad
c_\ell=8\cdot 2^\ell.
\end{equation}
The carrier length is halved while its width is doubled, so its total number
of scalar coefficients remains $8T$ at every level.

\subsection{Learned Lossless Subband Analysis}
\label{sec:lossless-analysis}

Each analysis stage applies a learned lifting transform to adjacent carrier
vectors. For level $\ell$, we first separate the carrier into even and odd
positions:
\[
\mathbf{e}^{(\ell)}_j=\mathbf{K}^{(\ell)}_{2j},
\qquad
\mathbf{o}^{(\ell)}_j=\mathbf{K}^{(\ell)}_{2j+1}.
\]
A learned predictor $\mathcal{P}_\ell$ estimates the odd vector from the even
vector, producing a detail coefficient. A learned update function
$\mathcal{U}_\ell$ then modifies the even vector using this detail:
\begin{equation}
\label{eq:lifting-analysis}
\begin{aligned}
\boldsymbol{\Delta}^{(\ell+1)}_j
&=\mathbf{o}^{(\ell)}_j
-\mathcal{P}_\ell\!\left(\mathbf{e}^{(\ell)}_j\right),
\mathbf{C}^{(\ell+1)}_j
=\mathbf{e}^{(\ell)}_j
+\mathcal{U}_\ell\!\left(\boldsymbol{\Delta}^{(\ell+1)}_j\right),\\
\mathbf{K}^{(\ell+1)}_j
&=\operatorname{cat}\!\left[
\mathbf{C}^{(\ell+1)}_j;
\boldsymbol{\Delta}^{(\ell+1)}_j
\right].
\end{aligned}
\end{equation}
Both $\mathcal{P}_\ell$ and $\mathcal{U}_\ell$ are pointwise low-rank MLPs:
RMSNorm, a rank-$r_{\mathrm{lift}}$ projection, a SiLU nonlinearity, and a
projection back to $c_\ell$ dimensions. Their final projections are
zero-initialized, so the carrier initially implements the lazy transform
$\boldsymbol{\Delta}=\mathbf{o}$ and $\mathbf{C}=\mathbf{e}$.

For fixed learned functions $\mathcal{P}_\ell$ and $\mathcal{U}_\ell$,
Equation~\eqref{eq:lifting-analysis} has the exact algebraic inverse
\begin{equation}
\label{eq:lifting-synthesis}
\begin{aligned}
\widehat{\mathbf{e}}^{(\ell)}_j
&=\mathbf{C}^{(\ell+1)}_j
-\mathcal{U}_\ell\!\left(
\boldsymbol{\Delta}^{(\ell+1)}_j
\right),
\widehat{\mathbf{o}}^{(\ell)}_j
=\boldsymbol{\Delta}^{(\ell+1)}_j
+\mathcal{P}_\ell\!\left(
\widehat{\mathbf{e}}^{(\ell)}_j
\right),\\
\widehat{\mathbf{K}}^{(\ell)}_{2j}
&=\widehat{\mathbf{e}}^{(\ell)}_j,
\qquad
\widehat{\mathbf{K}}^{(\ell)}_{2j+1}
=\widehat{\mathbf{o}}^{(\ell)}_j.
\end{aligned}
\end{equation}
Substituting Equation~\eqref{eq:lifting-analysis} into
Equation~\eqref{eq:lifting-synthesis} gives
$\widehat{\mathbf{e}}^{(\ell)}_j=\mathbf{e}^{(\ell)}_j$ and
$\widehat{\mathbf{o}}^{(\ell)}_j=\mathbf{o}^{(\ell)}_j$. Recursively applying
the inverse stages therefore reconstructs the original fixed-bit carrier
$\mathbf{K}^{(0)}$ exactly in real arithmetic. The detail coefficients are
also retained explicitly in the lossless carrier hierarchy.

\subsection{Semantic Analysis Path}
\label{sec:semantic-analysis}

The semantic stream is not constrained to be invertible. Instead, it learns a
compressed predictive representation. Adjacent semantic vectors are
concatenated and projected into the width of the next level. The projected
lossless carrier is injected through a learned low-rank map:
\begin{equation}
\label{eq:semantic-analysis}
\begin{aligned}
\mathbf{A}^{(\ell+1)}_j
&=
\mathbf{W}^{\mathrm{pair}}_\ell
\operatorname{cat}\!\left[
\mathbf{S}^{(\ell)}_{2j};
\mathbf{S}^{(\ell)}_{2j+1}
\right]
+
\alpha^{\mathrm{car}}_\ell
\mathcal{Q}^{\mathrm{car}}_\ell
\!\left(\mathbf{K}^{(\ell+1)}_j\right),\\
\mathbf{S}^{(\ell+1)}
&=
\mathcal{F}^{\downarrow}_\ell
\!\left(
\operatorname{RMSNorm}
\left(\mathbf{A}^{(\ell+1)}\right)
\right).
\end{aligned}
\end{equation}
Here, $\mathcal{Q}^{\mathrm{car}}_\ell$ is a low-rank carrier projection,
$\alpha^{\mathrm{car}}_\ell$ is a learned scalar, and
$\mathcal{F}^{\downarrow}_\ell$ is a stack of causal memory blocks.

A level-$\ell+1$ token with index $j$ summarizes two level-$\ell$ tokens and
therefore has a byte-level receptive-field endpoint
$(j+1)2^{\ell+1}-1$. Since the sequence mixers inside
$\mathcal{F}^{\downarrow}_\ell$ are causal, contextual processing may combine
the current parent with earlier parents but never with a parent whose byte span
ends later.

\subsection{Fast-Weight Memory Encoder}
\label{sec:fastweight-encoder}

The initial byte-resolution blocks and all down-path blocks use a gated
fast-weight recurrence \citep{schmidhuber1992fastweights,
schlag2021linear,yang2024gla}. For each head, projected query, key, and value
vectors are obtained from a causal depthwise convolution followed by a SiLU
activation. Queries and keys are normalized independently. A scalar
input-dependent log-decay $g_t\leq 0$ is produced for each head, with
$\gamma_t=\exp(g_t)\in(0,1]$.

The fast-weight state $\mathbf{M}_t$ and its readout are
\begin{equation}
\label{eq:fastweight-recurrence}
\mathbf{M}_t
=
\gamma_t\mathbf{M}_{t-1}
+
\mathbf{k}_t\mathbf{v}_t^{\top},
\qquad
\mathbf{o}_t
=
\mathbf{M}_t^{\top}\mathbf{q}_t.
\end{equation}
The outer-product update writes a key--value association into the recurrent
state, while the query retrieves a content-dependent combination of stored
values. Unlike a growing attention key--value cache, the recurrent state size
does not increase with sequence length.

Each encoder block uses pre-normalized residual mixing followed by a SwiGLU
feed-forward network:
\begin{equation}
\label{eq:fastweight-block}
\begin{aligned}
\widetilde{\mathbf{H}}
&=
\mathbf{H}
+
\mathcal{W}_{o}
\mathcal{R}
\!\left(
\operatorname{RMSNorm}(\mathbf{H})
\right),\\
\operatorname{FWBlock}(\mathbf{H})
&=
\widetilde{\mathbf{H}}
+
\operatorname{SwiGLU}
\!\left(
\operatorname{RMSNorm}(\widetilde{\mathbf{H}})
\right),
\end{aligned}
\end{equation}
where $\mathcal{R}$ denotes the recurrence in
Equation~\eqref{eq:fastweight-recurrence}. For full-sequence training, the
recurrence is evaluated using an exact chunk-parallel formulation. The
sequence is partitioned into fixed-size chunks, with intra-chunk interactions
computed by causal matrix multiplication and inter-chunk states propagated
through an equivalent lower-triangular decay computation. Chunking changes
the execution schedule but not the recurrence defined in
Equation~\eqref{eq:fastweight-recurrence}.

\subsection{Shared-State Bottleneck Refinement}
\label{sec:bottleneck-reasoning}

At the deepest level, the semantic stream contains 16 positions of width 128.
A shared FastWeight mixer iteratively refines a separate latent state while
holding the encoder bottleneck representation fixed. Let
$\mathbf{B}=\mathbf{S}^{(L)}$ and initialize $\mathbf{Z}_0=\mathbf{0}$.
Two recurrent updates are applied:
\begin{equation}
\label{eq:shared-state-reasoning}
\begin{aligned}
\mathbf{Z}_{r+1}
&=
\mathbf{Z}_{r}
+
\mathcal{R}_{\mathrm{bot}}\!\left(
\operatorname{RMSNorm}(\mathbf{B}+\mathbf{Z}_{r})
\right), \quad r\in\{0,1\},\\
\widetilde{\mathbf{S}}^{(L)}
&=\mathbf{B}+\mathbf{Z}_{2}.
\end{aligned}
\end{equation}
The same normalization and FastWeight mixer parameters are used in both
updates. The bottleneck module contains no additional feed-forward block and
introduces no iteration-specific parameters. Each update uses the causal
recurrence in Equation~\eqref{eq:fastweight-recurrence}, so the refined
bottleneck preserves completed-span causality. The decoder is initialized
from $\widetilde{\mathbf{S}}^{(L)}$.

\subsection{Causal Super-Resolution Decoder}
\label{sec:super-resolution}

The decoder progressively restores semantic representations from the refined
bottleneck at level $L$ to level $0$. At transition $\ell+1\rightarrow\ell$, a linear projection maps
each parent vector into two child proposals:
\begin{equation}
\label{eq:parent-proposals}
\left[
\mathbf{p}^{L,(\ell)}_j;
\mathbf{p}^{R,(\ell)}_j
\right]
=
\mathbf{W}^{\mathrm{up}}_\ell
\mathbf{Y}^{(\ell+1)}_j,
\end{equation}
where $\mathbf{Y}^{(\ell+1)}$ is the current decoded parent sequence.

A naive assignment of
$\mathbf{p}^{L,(\ell)}_j$ to position $2j$ would violate causality.
Parent $j$ summarizes both children $2j$ and $2j+1$, so it is only valid once
position $2j+1$ has entered the context. We therefore apply a shifted
unpacking rule:
\begin{equation}
\label{eq:causal-unpack}
\mathbf{U}^{(\ell)}_{2j}
=
\begin{cases}
\mathbf{0}, & j=0,\\
\mathbf{p}^{L,(\ell)}_{j-1}, & j>0,
\end{cases}
\qquad
\mathbf{U}^{(\ell)}_{2j+1}
=
\mathbf{p}^{R,(\ell)}_j.
\end{equation}
The right-child proposal from parent $j$ is used at position $2j+1$, where its
complete receptive field is available. The left-child position $2j$ instead
receives a proposal from parent $j-1$, whose receptive field ends before
position $2j$.

The shifted parent representation is fused with the encoder skip and the
carrier at the child resolution. Let
\[
\mathbf{Z}^{(\ell)}_t
=
\operatorname{cat}\!\left[
\mathbf{S}^{(\ell)}_t;
\mathbf{U}^{(\ell)}_t;
\alpha^{\mathrm{up}}_\ell
\mathcal{Q}^{\mathrm{up}}_\ell
\!\left(\mathbf{K}^{(\ell)}_t\right)
\right].
\]
The fused representation is
{\small
\begin{equation}
\label{eq:synthesis-fusion}
\begin{aligned}
\mathbf{G}^{(\ell)}_t
&=
\sigma\!\left(
\mathbf{W}^{g}_\ell\mathbf{Z}^{(\ell)}_t
\right),\\
\widehat{\mathbf{Y}}^{(\ell)}_t
&=
\operatorname{RMSNorm}
\left(
\mathbf{S}^{(\ell)}_t
+
\mathbf{G}^{(\ell)}_t
\odot
\mathbf{W}^{m}_\ell\mathbf{Z}^{(\ell)}_t
\right),\\
\mathbf{Y}^{(\ell)}
&=
\mathcal{F}^{\uparrow}_\ell
\!\left(
\widehat{\mathbf{Y}}^{(\ell)}
\right).
\end{aligned}
\end{equation}
}
The gate allows each position and channel to control how strongly decoded
parent context and carrier information modify the encoder skip.
$\mathcal{F}^{\uparrow}_\ell$ is a stack of causal self-attention blocks with
rotary position embeddings \citep{vaswani2017attention}. These blocks refine
the restored representation at the current resolution before the next
upsampling stage.

The decoder performs super-resolution over hidden representations rather than
predicting several unknown future bytes simultaneously. Every restored
position remains conditioned only on bytes that precede or equal that
position.

\subsection{Byte Prediction and Training Objective}
\label{sec:prediction-objective}

After super-resolution, the byte-resolution sequence
$\mathbf{H}^{(0)}=\mathbf{Y}^{(0)}$ is processed by a final stack of causal self-attention blocks. The output is
normalized and projected to the 256-byte vocabulary:
\begin{equation}
\label{eq:output-distribution}
\begin{aligned}
\mathbf{H}^{\mathrm{out}}
=
\mathcal{F}^{\mathrm{final}}
\left(
\mathbf{H}^{(0)}
\right),
\mathbf{z}_t
&=
\mathbf{W}^{\mathrm{vocab}}
\operatorname{RMSNorm}
\left(
\mathbf{H}^{\mathrm{out}}_t
\right),\\
p_\theta(x_{t+1}\mid x_{\leq t})
&=
\operatorname{softmax}
\left(
\mathbf{z}_t
\right)_{x_{t+1}}.
\end{aligned}
\end{equation}

The model is trained with next-byte negative log-likelihood:
\begin{equation}
\label{eq:training-loss}
\mathcal{L}_{\mathrm{NLL}}
=
-\frac{1}{T-1}
\sum_{t=0}^{T-2}
\log p_\theta
\left(
x_{t+1}\mid x_{\leq t}
\right).
\end{equation}
For evaluation, the average loss in nats is converted to bits per byte by
dividing by $\log 2$.

\section{Experiments}
\label{sec:experiments}

\subsection{Experimental Setup}
\label{sec:experimental-setup}

We train all models from scratch on the training split of FineWeb-Edu, represented directly as UTF-8 bytes with a vocabulary of 256 symbols. All runs use a context length of 2{,}048 bytes across three random seeds per model, and are matched to approximately 100 million trainable parameters. Unless otherwise stated, models are trained for 50{,}000 optimizer steps with a per-GPU batch size of 8 and gradient accumulation over 8 iterations. With four GPUs, this gives an effective batch size of 256 sequences, or 524{,}288 bytes per optimizer step, corresponding to approximately 26.2 billion byte presentations over training.

We use AdamW with $\beta_1=0.9$, $\beta_2=0.95$, weight decay $0.1$, and gradient clipping at $1.0$. The learning rate is linearly warmed up from zero to $3\times 10^{-4}$ over 2{,}000 steps and then cosine-decayed to $3\times 10^{-5}$. Training uses FP16 mixed precision, no dropout, and a streaming-data shuffle buffer of 20{,}000 examples. All models are trained and evaluated on four NVIDIA Tesla V100-SXM2 GPUs with 32\,GB of memory each.

\subsection{Baselines}
\label{sec:baselines}

We compare against four approximately 100M-parameter byte-level baselines:

\begin{itemize}
\item \textbf{Transformer:} a flat causal Transformer operating on all 2{,}048 byte positions at every layer \citep{vaswani2017attention};
\item \textbf{FastWeight:} a flat model replacing self-attention with gated fast-weight memory blocks \citep{schlag2021linear,yang2024gla};
\item \textbf{BLT:} the Byte Latent Transformer, which compresses bytes into entropy-dependent latent patches \citep{pagnoni2025blt}; and
\item \textbf{AU-Net:} an autoregressive U-Net with hierarchical downsampling, upsampling, and skip connections \citep{videau2025aunet}.
\end{itemize}

All baselines use the same training corpus, byte context length, optimizer-step budget, effective batch size, learning-rate schedule, numerical precision, and hardware. Parameter counts are matched as closely as permitted by each architecture.

\subsection{Reference Architecture}
\label{sec:reference-architecture}

Our reference model uses one initial fast-weight block, seven lossless analysis transitions, a two-step shared-state bottleneck refinement module, seven super-resolution synthesis transitions, and one final attention block. Starting from a 2{,}048-byte sequence, the hierarchy uses lengths
\[
(2048,1024,512,256,128,64,32,16)
\]
and semantic widths
\[
(1024,512,256,128,128,128,128,128).
\]
The semantic representation is therefore compressed rapidly at high resolution and held at width 128 throughout the deepest levels. The corresponding lossless carrier widths are
\[
(8,16,32,64,128,256,512,1024),
\]
preserving a constant total of $8T$ carrier coefficients at every resolution.

The initial and analysis-path mixers are gated fast-weight blocks, while all synthesis-path refinement blocks and the final block use causal self-attention with rotary position embeddings. The analysis and synthesis depths are both
\[
(2,2,2,3,3,3,4),
\]
listed from the highest-resolution transition to the deepest transition. Attention and fast-weight blocks use at most 16 heads. The byte-resolution model width is 1{,}024, with 16 heads.

Standard byte-resolution blocks use an MLP ratio of 4.0, while analysis and synthesis blocks use a ratio of 6.0. The lifting predictor and update functions use rank 64, as do carrier-to-semantic and detail projections. The fixed carrier dimension is 8, the minimum hierarchy length is 16, and direct all-scale fusion is disabled. The carrier residual gains are initialized to $0.1$. The shared-state bottleneck refinement module uses one 128-dimensional FastWeight mixer and two iterations, with parameters shared across both updates. The architecture has 100{,}158{,}582 trainable parameters, including 69{,}280 in the bottleneck refinement module. Fast-weight blocks use chunks of 64 positions and a causal depthwise-convolution kernel of size 4. Attention blocks use a rotary base of $10{,}000$. 



\subsection{Evaluation}
\label{sec:evaluation}

We evaluate language modeling using bits per byte (BPB), where lower is better. General-domain performance is measured on FineWeb-Edu, long-form modeling on the PG19 test set, and code modeling on Stack-smol. Multilingual performance is evaluated on English, German, Arabic, Hindi, Bengali, Thai, Vietnamese, and Swahili text from FLORES, and we report both language-level and macro-average BPB.

We additionally evaluate zero-shot multiple-choice accuracy on HellaSwag, PIQA, ARC-Easy, and ARC-Challenge. Robustness is measured using HellaSwag variants containing character deletion, repetition, Antspeak spacing, uppercasing, and random casing. For multiple-choice tasks, each candidate is scored by its conditional byte-level log-likelihood.

Evaluation efficiency is reported as aggregate processed bytes per second for configurations with available measurements. These measurements use the same evaluation implementation, batch construction, hardware, and numerical precision.


\section{Results}
\label{sec:results}

\subsection{Language Modeling Quality}
\label{sec:results-bpb}

Table~\ref{tab:bpb-results} summarizes bits per byte (BPB) on general-domain text, long-form text, code, and multilingual evaluation. Lower BPB indicates better byte-level language modeling. \textsc{DeepMem} achieves the strongest overall BPB results among the compared 100M-parameter systems. On FineWeb-Edu, it reaches \textbf{1.0134} BPB, improving over the matched Transformer at 1.0457, flat FastWeight at 1.0187, BLT at 1.0336, and AU-Net at 1.0319. On PG19, our model again performs best, achieving \textbf{1.4402} BPB and improving over the Transformer baseline at 1.4635. These results indicate that performing substantial computation at compressed resolutions can improve both in-domain and language modeling on FineWeb-Edu and PG19
under the fixed 2,048-byte context setting.

Code modeling shows a different pattern. On Stack-smol, the flat Transformer achieves the lowest Stack-smol
BPB among the primary models in
Table~\ref{tab:bpb-results} at \textbf{2.4474} BPB, while our model attains 2.4854 BPB. Even so, our method remains substantially stronger than flat FastWeight and BLT, suggesting that the proposed hierarchy preserves much of the benefit of flat modeling while trading off some performance on code. On the multilingual FLORES macro-average reported in the final column of Table~\ref{tab:bpb-results}, our model again performs best, reaching \textbf{1.9694} BPB. An additional all-scale fusion configuration
achieves 2.4082 BPB on Stack-smol
(Appendix~\ref{app:ablation-bpb}),
highlighting a task-specific architectural tradeoff.

\begin{table}[t]
\centering
\caption{Language-modeling performance in bits per byte (BPB). Lower is better. \textsc{DeepMem} (NR) represents \textsc{DeepMem} with no shared state refinement.}
\label{tab:bpb-results}

\begin{tabular}{lrrrr}
\toprule
\textbf{Model}
& \textbf{FineWeb-Edu}
& \textbf{PG19}
& \textbf{Stack-smol}
& \textbf{FLORES} \\
\midrule
Transformer
& 1.0457 $\pm$ 0.019
& 1.4635 $\pm$ 0.023
& \textbf{2.4474} $\pm$ 0.077
& 2.0082 $\pm$ 0.028 \\
FastWeight
& 1.0187 $\pm$ 0.008
& 1.4956 $\pm$ 0.019
& 2.9558 $\pm$ 0.047
& 2.0599 $\pm$ 0.022 \\
BLT
& 1.0336 $\pm$ 0.009
& 1.5920 $\pm$ 0.050
& 2.6138 $\pm$ 0.065
& 2.0233 $\pm$ 0.021 \\
AU-Net
& 1.0319 $\pm$ 0.006
& 1.4885 $\pm$ 0.022
& 2.4894 $\pm$ 0.058
& 2.0077 $\pm$ 0.020 \\
\hline
\textsc{DeepMem} (NR)
& 1.0145 $\pm$ 0.010
& 1.4462 $\pm$ 0.022
& 2.5021 $\pm$ 0.046
& 1.9758 $\pm$ 0.017 \\
\textsc{DeepMem}
& \textbf{1.0134} $\pm$ 0.008
& \textbf{1.4402} $\pm$ 0.017
& 2.4854 $\pm$ 0.034
& \textbf{1.9694} $\pm$ 0.020 \\
\bottomrule
\end{tabular}

\end{table}

\subsection{Multilingual Generalization}
\label{sec:results-multilingual}

Table~\ref{tab:flores-results} reports language-level BPB on eight FLORES languages. Our model obtains the best macro-average, \textbf{1.9694}, improving over the Transformer at 2.0082, BLT at 2.0233, FastWeight at 2.0599, and the AU-Net at 2.0077. The gains are distributed across multiple languages rather than concentrated in a single case. In particular, our model achieves the lowest BPB
on English, German, Arabic, and Thai. The variant
without shared-state refinement achieves the lowest
BPB on Bengali and Vietnamese.

The largest improvements over the Transformer appear in Thai, Arabic, and Vietnamese, where byte-level modeling benefits from combining a fixed 256-symbol vocabulary with multiscale semantic processing. The Transformer performs best on Swahili, while
flat FastWeight achieves the lowest BPB on Hindi.
The variant without shared-state refinement
performs best on Bengali and Vietnamese. Nevertheless, the overall pattern favors the proposed lossless multiscale architecture, indicating that its gains extend across distinct scripts and language families.

\begin{table*}[t]
\centering
\caption{Language-level BPB on FLORES. Lower is better. Macro is the unweighted mean across the eight languages. T, FW, AU, DM (NR), and DM represent Transformer, FastWeight, AU-Net, \textsc{DeepMem} with no shared state refinement, and \textsc{DeepMem}, respectively.}
\label{tab:flores-results}
\small
\setlength{\tabcolsep}{3.5pt}
\resizebox{\textwidth}{!}{
\begin{tabular}{lrrrrrrrrr}
\toprule
\textbf{Model}
& \textbf{English}
& \textbf{German}
& \textbf{Arabic}
& \textbf{Hindi}
& \textbf{Bengali}
& \textbf{Thai}
& \textbf{Vietnamese}
& \textbf{Swahili}
& \textbf{Macro} \\
\midrule
T
& 1.2013 $\pm$ 0.020
& 2.4014 $\pm$ 0.036
& 2.1194 $\pm$ 0.028
& 1.4265 $\pm$ 0.013
& 1.6157 $\pm$ 0.020
& 1.7914 $\pm$ 0.021
& 2.3884 $\pm$ 0.032
& \textbf{3.1216} $\pm$ 0.053
& 2.0082 $\pm$ 0.028 \\
FW
& 1.2484 $\pm$ 0.013
& 2.3674 $\pm$ 0.019
& 2.1350 $\pm$ 0.026
& \textbf{1.4196} $\pm$ 0.011
& 1.6205 $\pm$ 0.023
& 1.7722 $\pm$ 0.016
& 2.4547 $\pm$ 0.022
& 3.4613 $\pm$ 0.044
& 2.0599 $\pm$ 0.022 \\
BLT
& 1.2543 $\pm$ 0.013
& 2.3916 $\pm$ 0.027
& 2.1316 $\pm$ 0.027
& 1.4430 $\pm$ 0.013
& 1.6065 $\pm$ 0.022
& 1.7627 $\pm$ 0.016
& 2.3588 $\pm$ 0.016
& 3.2376 $\pm$ 0.032
& 2.0233 $\pm$ 0.021 \\
AU
& 1.2208 $\pm$ 0.019
& 2.3831 $\pm$ 0.022
& 2.1172 $\pm$ 0.028
& 1.4202 $\pm$ 0.011
& 1.6175 $\pm$ 0.023
& 1.7526 $\pm$ 0.010
& 2.3635 $\pm$ 0.019
& 3.1865 $\pm$ 0.027
& 2.0077 $\pm$ 0.020 \\
\hline
\textsc{DM} (NR)
& 1.2134 $\pm$ 0.013
& 2.3473 $\pm$ 0.014
& 2.0406 $\pm$ 0.019
& 1.4518 $\pm$ 0.015
& \textbf{1.5996} $\pm$ 0.012
& 1.6569 $\pm$ 0.016
& \textbf{2.2988} $\pm$ 0.018
& 3.1978 $\pm$ 0.028
& 1.9758 $\pm$ 0.017 \\
\textsc{DM}
& \textbf{1.2007} $\pm$ 0.015
& \textbf{2.3408} $\pm$ 0.019
& \textbf{2.0383} $\pm$ 0.019
& 1.4403 $\pm$ 0.013
& 1.6132 $\pm$ 0.016
& \textbf{1.6535} $\pm$ 0.016
& 2.3059 $\pm$ 0.027
& 3.1624 $\pm$ 0.032
& \textbf{1.9694} $\pm$ 0.020 \\
\bottomrule
\end{tabular}
}
\end{table*}

\subsection{Evaluation Throughput}
\label{sec:results-throughput}

Table~\ref{tab:throughput-results} reports BPB evaluation throughput in processed bytes per second for the comparison configurations. \textsc{DeepMem} processes an average of 119{,}718 bytes/s, compared with 147{,}023 bytes/s for the flat Transformer. This corresponds to 81.4\% of Transformer throughput, demonstrating that the hierarchical architecture maintains competitive evaluation efficiency while incorporating multiscale computation and shared-state bottleneck refinement.

The recorded evaluations show that hierarchical computation offers competitive throughput relative to flat FastWeight, BLT, and AU-Net. \textsc{DeepMem} achieves higher average throughput than all three architectures, while remaining slower than the flat Transformer. These results indicate that the memory-based analysis path and attention-based super-resolution decoder can support efficient byte-level evaluation despite the additional recurrent computation at the compressed bottleneck.

\begin{table}[t]
\centering
\caption{Mean BPB evaluation throughput in processed bytes per second. Higher is better.}
\label{tab:throughput-results}

\begin{tabular}{lrrrr}
\toprule
\textbf{Model}
& \textbf{FineWeb-Edu}
& \textbf{PG19}
& \textbf{Stack-smol}
& \textbf{Average} \\
\midrule
Transformer
& \textbf{147,788}
& \textbf{146,590}
& \textbf{146,690}
& \textbf{147,023} \\
FastWeight
& 87,577
& 86,663
& 87,095
& 87,112 \\
BLT
& 109,302
& 109,792
& 109,932
& 109,675 \\
AU-Net
& 114,565
& 115,658
& 118,965
& 116,396 \\
\hline
\textsc{DeepMem} (NR)
& 137,963
& 136,905
& 139,287
& 138,051 \\
\textsc{DeepMem}
& 118,963
& 120,905
& 119,287
& 119,718 \\
\bottomrule
\end{tabular}

\end{table}

\subsection{Multiple-Choice Evaluation}
\label{sec:results-multiple-choice}

Table~\ref{tab:mc-results} presents zero-shot
multiple-choice accuracy. \textsc{DeepMem} achieves
the highest HellaSwag accuracy (30.30\%), the highest
ARC-Challenge accuracy (23.41\%), and the highest
unweighted mean accuracy (37.76\%) among the main
comparison models. On PIQA, \textsc{DeepMem} achieves
60.83\%, while the variant without shared-state
refinement reaches 61.81\%. The full model exceeds
the matched Transformer on all four tasks, although
several improvements are small relative to the
reported run-level variation.

Flat FastWeight remains strongest on ARC-Easy, where it reaches \textbf{37.72\%} compared with 36.49\% for our model. Even so, our model achieves the strongest average across the four tasks, indicating that the proposed hierarchy does not sacrifice general zero-shot performance.

\begin{table}[t]
\centering
\caption{Zero-shot multiple-choice accuracy (\%). Higher is better. Mean is the unweighted average across the four tasks. \textsc{DeepMem} (NR) represents \textsc{DeepMem} with no shared state refinement.}
\label{tab:mc-results}

\begin{tabular}{lrrrrr}
\toprule
\textbf{Model}
& \textbf{HellaSwag}
& \textbf{PIQA}
& \textbf{ARC-E}
& \textbf{ARC-C}
& \textbf{Mean} \\
\midrule
Transformer
& 28.81 $\pm$ 0.4
& 60.55 $\pm$ 0.5
& 35.96 $\pm$ 0.6
& 22.74 $\pm$ 0.5
& 37.02 $\pm$ 0.5 \\
FastWeight
& 29.95 $\pm$ 0.4
& 60.61 $\pm$ 0.5
& \textbf{37.72} $\pm$ 0.6
& 20.07 $\pm$ 0.3
& 37.09 $\pm$ 0.5 \\
BLT
& 28.80 $\pm$ 0.4
& 60.55 $\pm$ 0.5
& 33.16 $\pm$ 0.7
& 21.07 $\pm$ 0.4
& 35.90 $\pm$ 0.5 \\
AU-Net
& 28.64 $\pm$ 0.4
& 60.66 $\pm$ 0.5
& 34.67 $\pm$ 0.6
& 20.86 $\pm$ 0.4
& 36.21 $\pm$ 0.5 \\
\hline
\textsc{DeepMem} (NR)
& 29.87 $\pm$ 0.4
& \textbf{61.81} $\pm$ 0.7
& 36.32 $\pm$ 0.6
& 21.74 $\pm$ 0.3
& 37.44 $\pm$ 0.5 \\
\textsc{DeepMem}
& \textbf{30.30} $\pm$ 0.4
& 60.83 $\pm$ 0.7
& 36.49 $\pm$ 0.6
& \textbf{23.41} $\pm$ 0.3
& \textbf{37.76} $\pm$ 0.5 \\
\bottomrule
\end{tabular}

\end{table}

\subsection{Robustness to Byte-Level Perturbations}
\label{sec:results-robustness}

Table~\ref{tab:robustness-results} evaluates robustness on HellaSwag under five character-level perturbations. Our model achieves the strongest accuracy on clean inputs, under repeated characters and random casing, and remains close to the strongest baseline under deletion, Antspeak spacing, and uppercasing. Its mean score across the clean and perturbed settings is 29.00\%, above the flat FastWeight model at 28.87\% and the Transformer at 27.73\%.

The pattern of robustness is informative. The strongest gains over the Transformer appear under random casing, deletion, and repetition, which perturb local byte patterns without necessarily changing the underlying semantics. This behavior is consistent with the intended role of the architecture: the lossless carrier and detail sidebands preserve fine-grained evidence, while the semantic hierarchy captures longer-range structure. Our model combines the highest average robustness with stronger BPB and stronger mean multiple-choice accuracy.

\begin{table}[t]
\centering
\caption{HellaSwag accuracy (\%) under byte-level perturbations. Higher is better. T, FW, AU and DM represent Transformer, FastWeight, AU-Net and \textsc{DeepMem}, respectively.}
\label{tab:robustness-results}
\small
\setlength{\tabcolsep}{4pt}

\begin{tabular}{lrrrrrrr}
\toprule
\textbf{Model}
& \textbf{Clean}
& \textbf{Drop}
& \textbf{Repeat}
& \textbf{Antspeak}
& \textbf{Uppercase}
& \textbf{Randomcase} \\
\midrule
T
& 28.81 $\pm$ 0.4
& 27.97 $\pm$ 0.5
& 27.50 $\pm$ 0.4
& 25.93 $\pm$ 0.4
& 28.79 $\pm$ 0.5
& 27.38 $\pm$ 0.4\\
FW
& 29.95 $\pm$ 0.4
& \textbf{29.18} $\pm$ 0.4
& 28.54 $\pm$ 0.4
& \textbf{27.06} $\pm$ 0.4
& \textbf{29.89} $\pm$ 0.4
& 28.58 $\pm$ 0.4\\
BLT
& 28.80 $\pm$ 0.4
& 28.98 $\pm$ 0.4
& 27.92 $\pm$ 0.4
& 26.02 $\pm$ 0.4
& 28.68 $\pm$ 0.4
& 27.42 $\pm$ 0.4\\
AU
& 28.64 $\pm$ 0.4
& 28.86 $\pm$ 0.4
& 27.54 $\pm$ 0.4
& 26.10 $\pm$ 0.4
& 28.42 $\pm$ 0.4
& 27.56 $\pm$ 0.4\\
\textsc{DM}
& \textbf{30.30} $\pm$ 0.4
& 29.16 $\pm$ 0.4
& \textbf{28.64} $\pm$ 0.4
& 26.89 $\pm$ 0.4
& 29.82 $\pm$ 0.3
& \textbf{29.16} $\pm$ 0.4\\
\bottomrule
\end{tabular}

\end{table}

\subsection{Overall Tradeoff}
\label{sec:results-summary}

Across the evaluated tasks, the proposed architecture achieves the best FineWeb-Edu BPB, the best PG19 BPB, the best FLORES macro-average BPB, and the highest mean multiple-choice and robustness accuracy among the main comparison models. The main weakness is code modeling, where the flat Transformer remains stronger. Evaluation throughput of the shared-state architecture reaches 119{,}718 bytes/s.

Taken together, the results support the central claim of the paper: exact lossless multiscale processing combined with shared-state bottleneck refinement can improve byte-level language modeling quality and multilingual generalization. Some task-specific tradeoffs remain, especially on code and ARC-Easy, but the overall pattern is favorable to the proposed approach.

\section{Discussion and Limitations}
\label{sec:discussion}

\paragraph{Discussion.}
Our results suggest that byte-level language modeling benefits from separating semantic abstraction from exact local information. The semantic hierarchy performs increasingly deep computation on shorter sequences, while the lossless carrier and detail sidebands preserve byte identity outside the semantic bottleneck. A shared-state refinement module iteratively refines the deepest semantic representation without duplicating mixer parameters. This division appears to improve natural-language and multilingual BPB. The comparative configurations (Supplementary) show that auxiliary carrier injection and direct all-scale fusion have task-dependent effects.

The architecture also highlights the importance of causal synthesis in hierarchical autoregressive models. Coarse representations summarize completed spans and therefore cannot be assigned directly to earlier positions within those spans. Delayed parent proposals allow the decoder to recover high-resolution representations without exposing future bytes. The bottleneck updates likewise preserve causal dependencies because they use a shared causal FastWeight mixer. More broadly, this provides a way to combine deep low-resolution computation with valid next-byte prediction.

The resulting performance suggests that hierarchical byte models need not choose between aggressive compression and local fidelity. In our model, the compressed semantic path can allocate capacity to longer-range structure because exact byte information remains available through the carrier hierarchy. Meanwhile, attention is concentrated in the synthesis path, where it refines representations after much of the computation has occurred at shorter resolutions. The comparative configurations (Supplementary) suggest that direct all-scale fusion can benefit individual languages, whereas omitting it allows shared-state refinement at the deepest bottleneck without a direct fusion path.

\paragraph{Limitations.}
Our study is limited to approximately 100M-parameter models, and a single training budget; larger-scale experiments are therefore needed. The model also trails the Transformer on Stack-smol, and its measured evaluation throughput is lower than that of the flat Transformer. Finally, exact invertibility holds in real arithmetic, while finite precision may introduce small reconstruction errors, and the fixed dyadic hierarchy is less flexible than learned variable-length segmentation.

\section{Conclusion}
\label{sec:conclusion}

We introduced a causal lossless subband architecture for byte-level language modeling that separates semantic compression from exact byte preservation. A FastWeight encoder performs deep computation over progressively shorter sequences, and a two-step shared-state bottleneck refinement module refines the compressed semantics, while an invertible carrier hierarchy, explicit detail sidebands, and a causal super-resolution decoder restore rich byte-resolution representations without future leakage.

At approximately 100M parameters, the model improves FineWeb-Edu, PG19, and multilingual BPB over a matched byte-level Transformer. It also achieves strong multiple-choice accuracy and robustness to byte-level perturbations. These results suggest that lossless multiscale representations with shared-state bottleneck refinement offer a practical alternative to flat byte processing, enabling tokenizer-free language models to combine strong predictive quality with exact byte preservation.

\paragraph{Code availability.} The code is available online at \url{https://github.com/akhadangi/deepmem}.


\newpage
{\small
\bibliographystyle{plain}
\bibliography{references}

@inproceedings{vaswani2017attention,
  title     = {Attention Is All You Need},
  author    = {Vaswani, Ashish and Shazeer, Noam and Parmar, Niki and
               Uszkoreit, Jakob and Jones, Llion and Gomez, Aidan N. and
               Kaiser, Lukasz and Polosukhin, Illia},
  booktitle = {Advances in Neural Information Processing Systems},
  volume    = {30},
  pages     = {5998--6008},
  year      = {2017}
}

@inproceedings{sennrich2016neural,
  title     = {Neural Machine Translation of Rare Words with Subword Units},
  author    = {Sennrich, Rico and Haddow, Barry and Birch, Alexandra},
  booktitle = {Proceedings of the 54th Annual Meeting of the Association for
               Computational Linguistics (Volume 1: Long Papers)},
  pages     = {1715--1725},
  address   = {Berlin, Germany},
  publisher = {Association for Computational Linguistics},
  year      = {2016},
  doi       = {10.18653/v1/P16-1162},
  url       = {https://aclanthology.org/P16-1162/}
}

@inproceedings{kudo2018sentencepiece,
  title     = {{SentencePiece}: A Simple and Language Independent Subword
               Tokenizer and Detokenizer for Neural Text Processing},
  author    = {Kudo, Taku and Richardson, John},
  booktitle = {Proceedings of the 2018 Conference on Empirical Methods in
               Natural Language Processing: System Demonstrations},
  pages     = {66--71},
  address   = {Brussels, Belgium},
  publisher = {Association for Computational Linguistics},
  year      = {2018},
  doi       = {10.18653/v1/D18-2012},
  url       = {https://aclanthology.org/D18-2012/}
}

@inproceedings{rust2021tokenizer,
  title     = {How Good Is Your Tokenizer? On the Monolingual Performance of
               Multilingual Language Models},
  author    = {Rust, Phillip and Pfeiffer, Jonas and Vuli{\'c}, Ivan and
               Ruder, Sebastian and Gurevych, Iryna},
  booktitle = {Proceedings of the 59th Annual Meeting of the Association for
               Computational Linguistics and the 11th International Joint
               Conference on Natural Language Processing},
  pages     = {3118--3135},
  publisher = {Association for Computational Linguistics},
  year      = {2021},
  doi       = {10.18653/v1/2021.acl-long.243},
  url       = {https://aclanthology.org/2021.acl-long.243/}
}

@article{xue2022byt5,
  title   = {{ByT5}: Towards a Token-Free Future with Pre-trained
             Byte-to-Byte Models},
  author  = {Xue, Linting and Barua, Aditya and Constant, Noah and
             Al-Rfou, Rami and Narang, Sharan and Kale, Mihir and
             Roberts, Adam and Raffel, Colin},
  journal = {Transactions of the Association for Computational Linguistics},
  volume  = {10},
  pages   = {291--306},
  year    = {2022},
  doi     = {10.1162/tacl_a_00461},
  url     = {https://aclanthology.org/2022.tacl-1.17/}
}

@article{clark2022canine,
  title   = {{CANINE}: Pre-training an Efficient Tokenization-Free Encoder
             for Language Representation},
  author  = {Clark, Jonathan H. and Garrette, Dan and Turc, Iulia and
             Wieting, John},
  journal = {Transactions of the Association for Computational Linguistics},
  volume  = {10},
  pages   = {73--91},
  year    = {2022},
  doi     = {10.1162/tacl_a_00448},
  url     = {https://aclanthology.org/2022.tacl-1.5/}
}

@inproceedings{tay2022charformer,
  title     = {Charformer: Fast Character Transformers via Gradient-Based
               Subword Tokenization},
  author    = {Tay, Yi and Tran, Vinh Q. and Ruder, Sebastian and Gupta, Jai and
               Chung, Hyung Won and Bahri, Dara and Qin, Zhen and
               Baumgartner, Simon and Yu, Cong and Metzler, Donald},
  booktitle = {International Conference on Learning Representations},
  year      = {2022},
  url       = {https://openreview.net/forum?id=JtBRnrlOEFN}
}

@inproceedings{yu2023megabyte,
  title     = {{MEGABYTE}: Predicting Million-Byte Sequences with
               Multiscale Transformers},
  author    = {Yu, Lili and Simig, D{\'a}niel and Flaherty, Colin and
               Aghajanyan, Armen and Zettlemoyer, Luke and Lewis, Mike},
  booktitle = {Advances in Neural Information Processing Systems},
  volume    = {36},
  pages     = {78808--78823},
  year      = {2023},
  url       = {https://proceedings.neurips.cc/paper_files/paper/2023/hash/f8f78f8043f35890181a824e53a57134-Abstract-Conference.html}
}

@inproceedings{slagle2024spacebyte,
  title     = {{SpaceByte}: Towards Deleting Tokenization from Large
               Language Modeling},
  author    = {Slagle, Kevin},
  booktitle = {Advances in Neural Information Processing Systems},
  volume    = {37},
  year      = {2024},
  url       = {https://papers.nips.cc/paper_files/paper/2024/hash/e1f418450107c4a0ddc16d008d131573-Abstract-Conference.html}
}

@inproceedings{wang2024mambabyte,
  title     = {{MambaByte}: Token-Free Selective State Space Model},
  author    = {Wang, Junxiong and Gangavarapu, Tushaar and
               Yan, Jing Nathan and Rush, Alexander M.},
  booktitle = {Conference on Language Modeling},
  year      = {2024},
  eprint    = {2401.13660},
  archivePrefix = {arXiv},
  primaryClass  = {cs.CL},
  url       = {https://arxiv.org/abs/2401.13660}
}

@inproceedings{pagnoni2025blt,
  title     = {Byte Latent Transformer: Patches Scale Better Than Tokens},
  author    = {Pagnoni, Artidoro and Pasunuru, Ramakanth and
               Rodriguez, Pedro and Nguyen, John and Muller, Benjamin and
               Li, Margaret and Zhou, Chunting and Yu, Lili and
               Weston, Jason E. and Zettlemoyer, Luke and Ghosh, Gargi and
               Lewis, Mike and Holtzman, Ari and Iyer, Srini},
  booktitle = {Proceedings of the 63rd Annual Meeting of the Association for
               Computational Linguistics (Volume 1: Long Papers)},
  pages     = {9238--9258},
  address   = {Vienna, Austria},
  publisher = {Association for Computational Linguistics},
  year      = {2025},
  doi       = {10.18653/v1/2025.acl-long.453},
  url       = {https://aclanthology.org/2025.acl-long.453/}
}

@inproceedings{kallini2025mrt5,
  title     = {{MrT5}: Dynamic Token Merging for Efficient Byte-Level
               Language Models},
  author    = {Kallini, Julie and Murty, Shikhar and Manning, Christopher D. and
               Potts, Christopher and Csord{\'a}s, R{\'o}bert},
  booktitle = {International Conference on Learning Representations},
  year      = {2025},
  url       = {https://openreview.net/forum?id=VYWBMq1L7H}
}

@article{egli2025mblm,
  title   = {Multiscale Byte Language Models: A Hierarchical Architecture
             for Causal Million-Length Sequence Modeling},
  author  = {Egli, Eric and Manica, Matteo and Born, Jannis},
  journal = {arXiv preprint arXiv:2502.14553},
  year    = {2025},
  eprint  = {2502.14553},
  archivePrefix = {arXiv},
  primaryClass  = {cs.LG},
  url     = {https://arxiv.org/abs/2502.14553}
}

@article{neitemeier2025hierarchical,
  title   = {Hierarchical Autoregressive Transformers: Combining Byte- and
             Word-Level Processing for Robust, Adaptable Language Models},
  author  = {Neitemeier, Pit and Deiseroth, Bj{\"o}rn and Eichenberg, Constantin and
             Balles, Lukas},
  journal = {arXiv preprint arXiv:2501.10322},
  year    = {2025},
  eprint  = {2501.10322},
  archivePrefix = {arXiv},
  primaryClass  = {cs.CL},
  url     = {https://arxiv.org/abs/2501.10322}
}

@inproceedings{dai2020funnel,
  title     = {Funnel-Transformer: Filtering out Sequential Redundancy for
               Efficient Language Processing},
  author    = {Dai, Zihang and Lai, Guokun and Yang, Yiming and Le, Quoc V.},
  booktitle = {Advances in Neural Information Processing Systems},
  volume    = {33},
  pages     = {4271--4282},
  year      = {2020},
  url       = {https://arxiv.org/abs/2006.03236}
}

@inproceedings{nawrot2022hourglass,
  title     = {Hierarchical Transformers Are More Efficient Language Models},
  author    = {Nawrot, Piotr and Tworkowski, Szymon and
               Tyrolski, Micha{\l} and Kaiser, Lukasz and Wu, Yuhuai and
               Szegedy, Christian and Michalewski, Henryk},
  booktitle = {Findings of the Association for Computational Linguistics:
               NAACL 2022},
  pages     = {1559--1571},
  address   = {Seattle, United States},
  publisher = {Association for Computational Linguistics},
  year      = {2022},
  doi       = {10.18653/v1/2022.findings-naacl.117},
  url       = {https://aclanthology.org/2022.findings-naacl.117/}
}

@inproceedings{videau2025aunet,
  title     = {From Bytes to Ideas: Language Modeling with Autoregressive
               {U}-Nets},
  author    = {Videau, Mathurin and Youbi Idrissi, Badr and Leite, Alessandro and
               Schoenauer, Marc and Teytaud, Olivier and Lopez-Paz, David},
  booktitle = {Advances in Neural Information Processing Systems},
  volume    = {38},
  year      = {2025},
  url       = {https://papers.nips.cc/paper_files/paper/2025/hash/871d547a8922ca600eea526a1bd40b2c-Abstract-Conference.html}
}

@article{kallini2026fastblt,
  title   = {Fast Byte Latent Transformer},
  author  = {Kallini, Julie and Pagnoni, Artidoro and Limisiewicz, Tomasz and
             Ghosh, Gargi and Zettlemoyer, Luke and Potts, Christopher and
             Han, Xiaochuang and Iyer, Srinivasan},
  journal = {arXiv preprint arXiv:2605.08044},
  year    = {2026},
  eprint  = {2605.08044},
  archivePrefix = {arXiv},
  primaryClass  = {cs.CL},
  doi     = {10.48550/arXiv.2605.08044},
  url     = {https://arxiv.org/abs/2605.08044}
}

@article{burt1983laplacian,
  title   = {The Laplacian Pyramid as a Compact Image Code},
  author  = {Burt, Peter J. and Adelson, Edward H.},
  journal = {IEEE Transactions on Communications},
  volume  = {31},
  number  = {4},
  pages   = {532--540},
  year    = {1983},
  doi     = {10.1109/TCOM.1983.1095851}
}

@article{sweldens1998lifting,
  title   = {The Lifting Scheme: A Construction of Second Generation Wavelets},
  author  = {Sweldens, Wim},
  journal = {SIAM Journal on Mathematical Analysis},
  volume  = {29},
  number  = {2},
  pages   = {511--546},
  year    = {1998},
  doi     = {10.1137/S0036141095289051}
}

@inproceedings{ronneberger2015unet,
  title     = {{U-Net}: Convolutional Networks for Biomedical Image Segmentation},
  author    = {Ronneberger, Olaf and Fischer, Philipp and Brox, Thomas},
  booktitle = {Medical Image Computing and Computer-Assisted Intervention},
  series    = {Lecture Notes in Computer Science},
  volume    = {9351},
  pages     = {234--241},
  publisher = {Springer},
  year      = {2015},
  doi       = {10.1007/978-3-319-24574-4_28}
}

@article{schmidhuber1992fastweights,
  title   = {Learning to Control Fast-Weight Memories: An Alternative to
             Dynamic Recurrent Networks},
  author  = {Schmidhuber, J{\"u}rgen},
  journal = {Neural Computation},
  volume  = {4},
  number  = {1},
  pages   = {131--139},
  year    = {1992},
  doi     = {10.1162/neco.1992.4.1.131}
}

@inproceedings{katharopoulos2020transformers,
  title     = {Transformers Are {RNN}s: Fast Autoregressive Transformers
               with Linear Attention},
  author    = {Katharopoulos, Angelos and Vyas, Apoorv and Pappas, Nikolaos and
               Fleuret, Fran{\c{c}}ois},
  booktitle = {Proceedings of the 37th International Conference on Machine Learning},
  series    = {Proceedings of Machine Learning Research},
  volume    = {119},
  pages     = {5156--5165},
  publisher = {PMLR},
  year      = {2020},
  url       = {https://proceedings.mlr.press/v119/katharopoulos20a.html}
}

@inproceedings{schlag2021linear,
  title     = {Linear Transformers Are Secretly Fast Weight Programmers},
  author    = {Schlag, Imanol and Irie, Kazuki and Schmidhuber, J{\"u}rgen},
  booktitle = {Proceedings of the 38th International Conference on Machine Learning},
  series    = {Proceedings of Machine Learning Research},
  volume    = {139},
  pages     = {9355--9366},
  publisher = {PMLR},
  year      = {2021},
  url       = {https://proceedings.mlr.press/v139/schlag21a.html}
}

@inproceedings{yang2024gla,
  title     = {Gated Linear Attention Transformers with Hardware-Efficient Training},
  author    = {Yang, Songlin and Wang, Bailin and Shen, Yikang and
               Panda, Rameswar and Kim, Yoon},
  booktitle = {Proceedings of the 41st International Conference on Machine Learning},
  series    = {Proceedings of Machine Learning Research},
  volume    = {235},
  pages     = {56501--56523},
  publisher = {PMLR},
  year      = {2024},
  url       = {https://proceedings.mlr.press/v235/yang24ab.html}
}
}

\newpage

\appendix
\renewcommand{\thefigure}{\thesection.\arabic{figure}}
\renewcommand{\thetable}{\thesection.\arabic{table}}
\renewcommand{\theequation}{\thesection.\arabic{equation}}
\setcounter{figure}{0}
\setcounter{table}{0}
\setcounter{equation}{0}

\section{Ablation Studies}
\label{app:ablations}

We study the FastWeight memory encoder, the injection of the lossless carrier
into the semantic analysis path, direct all-scale fusion of semantic and
detail features, and two-step shared-state bottleneck refinement in
\textsc{DeepMem}. The comparison configurations use the same FineWeb-Edu
training data, 2,048-byte context length, 50,000-step training budget,
optimizer schedule, effective batch size, and four NVIDIA V100-SXM2-32GB GPUs.
The all-scale fusion, no-carrier-injection, and all-attention configurations
have no shared-state bottleneck update; their respective comparisons isolate
the indicated analysis or fusion component. 
The full \textsc{DeepMem} model uses two
shared-state bottleneck updates without direct
all-scale fusion. The no-shared-state variant
removes these updates while retaining the same
analysis and synthesis architecture. The
all-scale fusion variant adds direct delayed
cross-scale feature fusion to the model without
shared-state updates. These configurations
allow us to examine bottleneck refinement and
fusion separately.

The
\emph{all-attention} configuration replaces the initial and down-path FastWeight
blocks with causal self-attention, while the synthesis and final paths use
attention. The \emph{no encoder carrier injection} configuration bypasses the
carrier-to-semantic projection in every analysis stage but retains the fixed
byte carrier, invertible lifting transform, carrier features in synthesis,
and detail-sideband fusion. The \emph{no all-scale fusion} configuration
retains the lossless carrier hierarchy and causal super-resolution decoder,
with no direct all-scale fusion or iterative bottleneck update.

\subsection{Language Modeling and Throughput}
\label{app:ablation-bpb}

Table~\ref{tab:ablation-bpb} reports language-modeling quality and
evaluation throughput. Comparing the configurations without shared-state
reasoning, replacing the FastWeight encoder with attention causes the clearest
and most consistent degradation. FineWeb-Edu BPB increases from $1.0145$ to
$1.0613$, a relative degradation of $4.6\%$. PG19 increases by $2.5\%$,
and FLORES macro BPB increases by $3.7\%$. The all-attention configuration
improves Stack-smol from $2.5021$ to $2.4527$ BPB and raises average
evaluation throughput by approximately $0.3\%$. This establishes a
quality--efficiency tradeoff among these comparison configurations.

Removing encoder-side carrier injection from the no-shared-state
configuration has only a small effect on aggregate BPB. FineWeb-Edu is
effectively unchanged, PG19 worsens by $0.0020$ BPB, Stack-smol improves by
$0.0176$, and FLORES macro worsens by $0.0072$. Because the carrier remains
available during synthesis and through detail fusion, this result should not
be interpreted as evidence that the lossless carrier itself is unnecessary.

Relative to the all-scale fusion configuration,
removing fusion changes FineWeb-Edu BPB from
1.0142 to 1.0145, PG19 BPB from 1.4413 to
1.4462, and Stack-smol BPB from 2.4082 to
2.5021. In contrast, FLORES macro BPB improves
from 1.9851 to 1.9758. These results suggest
that direct all-scale fusion benefits code
modeling in this comparison, whereas the
configuration without fusion performs better
on multilingual average BPB.

Comparing \textsc{DeepMem} with the variant
without shared-state refinement, FineWeb-Edu
BPB decreases from 1.0145 to 1.0134, PG19
from 1.4462 to 1.4402, Stack-smol from 2.5021
to 2.4854, and FLORES macro BPB from 1.9758
to 1.9694. Thus, the full model achieves
lower BPB on all four evaluations in this
comparison. However, average BPB-evaluation
throughput decreases from approximately
138,052 to 119,718 bytes/s. The improvements
in BPB are modest, and additional controlled
runs are needed to establish their statistical
reliability.

\begin{table*}[t]
\centering
\caption{Architecture results for language modeling and BPB-evaluation throughput.
BPB (mean) is lower better; processed bytes per second (B/s) is higher better.}
\label{tab:ablation-bpb}
\scriptsize
\setlength{\tabcolsep}{10pt}
\resizebox{\textwidth}{!}{
\begin{tabular}{lrrrrrrrr}
\toprule
& \multicolumn{4}{c}{\textbf{Mean BPB}}
& \multicolumn{4}{c}{\textbf{Throughput (B/s)}} \\
\cmidrule(lr){2-5}
\cmidrule(lr){6-9}
\textbf{Variant}
& \textbf{FineWeb}
& \textbf{PG19}
& \textbf{Stack}
& \textbf{FLORES}
& \textbf{FineWeb}
& \textbf{PG19}
& \textbf{Stack}
& \textbf{Average} \\
\midrule
\textsc{DeepMem}
& \textbf{1.0134}
& \textbf{1.4402}
& 2.4854
& \textbf{1.9694}
& 118,963
& 120,905
& 119,287
& 119,718 \\
No shared-state refinement
& 1.0145
& 1.4462
& 2.5021
& 1.9758
& 137,963
& \textbf{136,905}
& 139,287
& 138,051 \\
No carrier$\rightarrow$sem.
& 1.0141
& 1.4482
& 2.4845
& 1.9830
& 121,953
& 122,663
& 122,754
& 122,457 \\
All-scale fusion
& 1.0142
& 1.4413
& \textbf{2.4082}
& 1.9851
& 117,374
& 118,468
& 119,369
& 118,404 \\
All attention
& 1.0613
& 1.4827
& 2.4527
& 2.0498
& \textbf{140,812}
& 134,798
& \textbf{139,944}
& \textbf{138,518} \\
\bottomrule
\end{tabular}
}
\end{table*}

\subsection{Multilingual Effects}
\label{app:ablation-multilingual}

Table~\ref{tab:ablation-flores} provides
language-level FLORES results for the
architecture variants. \textsc{DeepMem}
achieves the lowest macro-average BPB and
the lowest values on English, German, Arabic,
Hindi, Thai, and Swahili among these variants.
The configuration without shared-state
refinement achieves the lowest values on
Bengali and Vietnamese.

Comparing the two configurations without
shared-state refinement, removing all-scale
fusion improves Thai BPB by 0.0318,
Vietnamese by 0.0268, Bengali by 0.0139,
and Arabic by 0.0124. In contrast, the
all-scale fusion configuration performs better
on Swahili by 0.0230 BPB. The macro-average
improves from 1.9851 to 1.9758 without
all-scale fusion. These comparisons are
exploratory because only one training run
is available for each configuration.

The all-attention configuration has higher
BPB than the all-scale fusion configuration
on all eight languages. The largest absolute
differences are observed on Vietnamese
(+0.1340 BPB), German (+0.0999), Thai
(+0.0822), and Arabic (+0.0661). Compared
with the configuration without shared-state
refinement, the full \textsc{DeepMem} model
improves FLORES macro BPB by 0.0064.
Whether these differences isolate individual
components depends on the verified
configuration settings.

\begin{table*}[t]
\centering
\caption{Mean FLORES BPB for the architecture configurations. Lower is better.}
\label{tab:ablation-flores}
\scriptsize
\setlength{\tabcolsep}{8pt}
\resizebox{\textwidth}{!}{
\begin{tabular}{lrrrrrrrrr}
\toprule
\textbf{Variant}
& \textbf{English}
& \textbf{German}
& \textbf{Arabic}
& \textbf{Hindi}
& \textbf{Bengali}
& \textbf{Thai}
& \textbf{Vietnamese}
& \textbf{Swahili}
& \textbf{Macro} \\
\midrule
\textsc{DeepMem}
& \textbf{1.2007}
& \textbf{2.3408}
& \textbf{2.0383}
& \textbf{1.4403}
& 1.6132
& \textbf{1.6535}
& 2.3059
& \textbf{3.1624}
& \textbf{1.9694} \\
No shared-state refinement
& 1.2134
& 2.3473
& 2.0406
& 1.4518
& \textbf{1.5996}
& 1.6569
& \textbf{2.2988}
& 3.1978
& 1.9758 \\
No carrier$\rightarrow$sem.
& 1.2206
& 2.3625
& 2.0522
& 1.4513
& 1.6225
& 1.6542
& 2.3284
& 3.1726
& 1.9830 \\
All-scale fusion
& 1.2187
& 2.3512
& 2.0530
& 1.4553
& 1.6135
& 1.6887
& 2.3256
& 3.1748
& 1.9851 \\
All attention
& 1.2246
& 2.4511
& 2.1191
& 1.5030
& 1.6574
& 1.7709
& 2.4596
& 3.2131
& 2.0498 \\
\bottomrule
\end{tabular}
}
\end{table*}

\subsection{Multiple-Choice Evaluation}
\label{app:ablation-multiple-choice}

Table~\ref{tab:ablation-mc} compares zero-shot multiple-choice accuracy.
\textsc{DeepMem} achieves the strongest HellaSwag accuracy at $30.30\%$
and the highest mean accuracy at $37.76\%$. The highest PIQA result among the reported
configurations is achieved by the variant
without shared-state refinement (61.81\%), while
the highest ARC-Challenge result is achieved by the no-carrier-injection configuration.
\textsc{DeepMem} also achieves the strongest ARC-Easy accuracy.

Removing encoder-side carrier injection
changes mean accuracy from 37.44\% to
37.53\% relative to the no-shared-state
configuration. The all-attention variant
achieves 35.95\% mean accuracy, compared
with 37.44\% for the no-shared-state variant.
However, these comparisons can only be
attributed to individual architectural
components when the remaining configuration
settings are identical.

\begin{table}[t]
\centering
\caption{Mean zero-shot multiple-choice accuracy (\%) for the architecture
configurations. Higher is better.}
\label{tab:ablation-mc}

\begin{tabular}{lrrrrr}
\toprule
\textbf{Variant}
& \textbf{Hella}
& \textbf{PIQA}
& \textbf{ARC-E}
& \textbf{ARC-C}
& \textbf{Mean} \\
\midrule
\textsc{DeepMem}
& \textbf{30.30}
& 60.83
& \textbf{36.49}
& 23.41
& \textbf{37.76} \\
No shared-state refinement
& 29.87
& \textbf{61.81}
& 36.32
& 21.74
& 37.44 \\
No carrier$\rightarrow$sem.
& 29.76
& 60.99
& 35.61
& \textbf{23.75}
& 37.53 \\
All attention
& 28.70
& 59.68
& 33.68
& 21.74
& 35.95 \\
\bottomrule
\end{tabular}

\end{table}

\subsection{Robustness to Byte-Level Perturbations}
\label{app:ablation-robustness}

Table~\ref{tab:ablation-robustness} reports HellaSwag accuracy when only
the prompt is perturbed. \textsc{DeepMem} achieves the strongest mean
accuracy across the clean and perturbed settings at $29.00\%$, with the best
clean, uppercase, and random-casing results. The no-carrier-injection
configuration has the strongest deletion and repetition scores, while the no-shared-state configuration
achieves the highest Antspeak accuracy (27.23\%).

Relative to the no-shared-state configuration, \textsc{DeepMem} improves
clean accuracy by $0.43$ points, uppercase by $0.32$ points, and random
casing by $0.38$ points, raising the overall mean by $0.15$ points.
Deletion and repetition accuracy also improve slightly, whereas Antspeak
accuracy decreases by $0.34$ points. 

The comparison results do not establish that either encoder-side carrier
injection or all-scale fusion is individually necessary for robustness.
Their effects are perturbation dependent and partially interchangeable
with the remaining carrier, skip, and synthesis paths. Replacing FastWeight with attention causes
a reduction under every perturbation relative to the no shared state configuration and lowers its average by $1.43$ points. The memory-based
analysis path therefore remains important for robustness, while
\textsc{DeepMem} achieves the strongest aggregate score among these
configurations.

\begin{table*}[t]
\centering
\caption{Mean HellaSwag accuracy (\%) under prompt-level byte perturbations.
Higher is better.}
\label{tab:ablation-robustness}
\small
\setlength{\tabcolsep}{5pt}
\begin{tabular}{lrrrrrrr}
\toprule
\textbf{Variant}
& \textbf{Clean}
& \textbf{Drop}
& \textbf{Repeat}
& \textbf{Antspeak}
& \textbf{Uppercase}
& \textbf{Randomcase}
& \textbf{Mean} \\
\midrule
\textsc{DeepMem}
& \textbf{30.30}
& 29.16
& 28.64
& 26.89
& \textbf{29.82}
& \textbf{29.16}
& \textbf{29.00} \\
No carrier$\rightarrow$sem.
& 29.76
& \textbf{29.33}
& \textbf{28.66}
& 26.90
& 29.44
& 28.52
& 28.77 \\
No shared-state refinement
& 29.87
& 29.13
& 28.56
& \textbf{27.23}
& 29.50
& 28.78
& 28.85 \\
All attention
& 28.70
& 27.56
& 26.95
& 25.92
& 28.07
& 27.32
& 27.42 \\
\bottomrule
\end{tabular}
\end{table*}

\subsection{Ablation Summary}
\label{app:ablation-summary}

The ablations indicate that the architectural
components exhibit distinct tradeoffs.
Removing shared-state bottleneck refinement
from \textsc{DeepMem} produces slightly higher
BPB on FineWeb-Edu, PG19, Stack-smol, and
FLORES, while improving dense evaluation
throughput.

Introducing direct all-scale fusion into
the configuration without shared-state
refinement improves PG19 and Stack-smol
BPB, while worsening FLORES macro-average
BPB. Encoder-side carrier injection has
smaller and task-dependent effects, although
the lossless carrier remains available
through other paths.

The all-attention configuration generally
shows weaker natural-language and
multilingual BPB than the FastWeight-based
comparison configurations. 

Overall, these comparisons suggest that
multiscale fusion, memory-based encoding,
and iterative bottleneck refinement
contribute differently across tasks.

\section{Causality Analysis and Verification}
\label{app:causality}

Hierarchical autoregressive models require causality not only within their
sequence mixers, but also across downsampling, upsampling, skip connections,
and any optional cross-scale fusion. In this section, we first establish a span-based
causality invariant for \textsc{DeepMem} and then report repeatability-controlled
future-perturbation tests of the shared hierarchical components.

\subsection{Span-Based Causality}
\label{app:causality-proof}

A representation at hierarchy level $\ell$ and index $j$ corresponds to a
completed byte span of length $2^\ell$. We define the final byte position of
that span as
\begin{equation}
\label{eq:causal-span-end}
e_\ell(j) = (j+1)2^\ell - 1.
\end{equation}
A level-$\ell$ representation $\mathbf{R}^{(\ell)}_j$ is causally valid if
it depends only on bytes up to this endpoint:
\begin{equation}
\label{eq:causal-span-invariant}
\mathbf{R}^{(\ell)}_j
=
f_{\ell,j}
\left(
x_0,\ldots,x_{e_\ell(j)}
\right).
\end{equation}
At level zero, the embedding and fixed carrier at position $t$ depend only on
$x_t$, so the invariant holds directly.

\paragraph{Lossless analysis.}
The analysis transition from level $\ell$ to $\ell+1$ combines child
positions $2j$ and $2j+1$. The later child has endpoint
\begin{equation}
\label{eq:causal-analysis-endpoint}
e_\ell(2j+1)
=
(2j+2)2^\ell-1
=
(j+1)2^{\ell+1}-1
=
e_{\ell+1}(j).
\end{equation}
Therefore, the semantic pair projection, learned carrier lifting transform,
and resulting detail coefficient depend only on bytes up to the endpoint of
parent $j$. The FastWeight encoder is also causal: at parent index $j$, its
recurrent state contains writes only from indices $i\leq j$. Hence the
semantic representation, coarse carrier, and detail sideband at every
analysis level satisfy Equation~\eqref{eq:causal-span-invariant}.

\paragraph{Shared-state bottleneck refinement.}
The two bottleneck updates apply the same causal FastWeight mixer and
pointwise normalization to the sum of the fixed encoder bottleneck and the
current latent state. Initially, the latent state is zero. By induction, each
update depends only on the current and preceding completed bottleneck spans,
so the refined bottleneck preserves Equation~\eqref{eq:causal-span-invariant}.

\paragraph{Causal super-resolution synthesis.}
A parent at level $\ell+1$ summarizes both of its level-$\ell$ children. Its
right-child proposal can safely be assigned to position $2j+1$, since
\begin{equation}
\label{eq:causal-right-child}
e_{\ell+1}(j)=e_\ell(2j+1).
\end{equation}
Assigning the same parent to the earlier child $2j$ would expose information
from child $2j+1$. The unpacking rule defined in the main paper instead assigns the
left-child position a proposal from parent $j-1$. For $j>0$, this parent ends
at
\begin{equation}
\label{eq:causal-left-child}
e_{\ell+1}(j-1)
=
j2^{\ell+1}-1
\leq
(2j+1)2^\ell-1
=
e_\ell(2j).
\end{equation}
The first left-child position receives a zero proposal. Thus, both branches of
the shifted unpacking operation are causally valid. The encoder skip and
carrier feature used at each child position satisfy the same endpoint
constraint, and the subsequent decoder refinement uses causally masked
self-attention.

\paragraph{All-scale fusion.}
A semantic or detail feature at level $\ell$ summarizes a span of
$s_\ell=2^\ell$ bytes. The delay-and-expand operator in this configuration repeats each feature to byte resolution and
shifts it by $s_\ell-1$ positions. Feature $j$ first appears at
\begin{equation}
\label{eq:causal-multiscale-endpoint}
j s_\ell + (s_\ell-1)
=
(j+1)2^\ell-1
=
e_\ell(j),
\end{equation}
which is exactly the endpoint of its source span. It may influence later byte
positions, but it cannot appear before all bytes in its source span are part
of the available context.

\paragraph{Final prediction.}
Pointwise projections and normalization do not enlarge the dependency set,
and the final byte-resolution attention block is causal. Consequently, the
logit vector at position $t$ satisfies
\begin{equation}
\label{eq:causal-logits}
\mathbf{z}_t
=
g_t(x_0,\ldots,x_t),
\qquad
p_\theta(x_{t+1}\mid\mathbf{x})
=
p_\theta(x_{t+1}\mid x_{\leq t}).
\end{equation}
Changing bytes after position $t$ therefore cannot alter any logit at
positions $0,\ldots,t$.

\subsection{Future-Perturbation Protocol}
\label{app:causality-protocol}

We verify the complete computation graph by comparing sequences that share a
prefix but differ at every later byte. For a cutoff $t$, let
$\widetilde{\mathbf{x}}^{(t)}$ satisfy
\begin{equation}
\label{eq:future-perturbation}
\widetilde{x}^{(t)}_i=x_i
\quad\text{for }i\leq t,
\qquad
\widetilde{x}^{(t)}_i\neq x_i
\quad\text{for }i>t.
\end{equation}
We measure the largest change in any shared-prefix logit:
\begin{equation}
\label{eq:causality-prefix-error}
\delta_t
=
\max_{\substack{i\leq t\\v\in\{0,\ldots,255\}}}
\left|
z_{i,v}(\mathbf{x})
-
z_{i,v}\!\left(\widetilde{\mathbf{x}}^{(t)}\right)
\right|.
\end{equation}
To distinguish future-perturbation effects from run-to-run numerical
variation, we also execute the unmodified input twice and compute
\begin{equation}
\label{eq:causality-repeatability}
\delta_{\mathrm{rep}}
=
\max_{i,v}
\left|
z_{i,v}^{(1)}(\mathbf{x})
-
z_{i,v}^{(2)}(\mathbf{x})
\right|.
\end{equation}
A test passes only if both $\delta_{\mathrm{rep}}$ and every prefix difference
are at most $\varepsilon=10^{-5}$.

The recorded tests use the 2,048-byte all-scale fusion comparison
configuration without shared-state bottleneck updates, with hierarchy lengths
\[
(2048,1024,512,256,128,64,32,16).
\]
They are executed on CUDA in FP32 with TF32 disabled. Flash and
memory-efficient scaled-dot-product attention are disabled in favor of the
mathematical attention implementation. The same-input repeatability control is
included in both tests.

The tested configuration is randomly initialized. This is sufficient to
exercise structural causality of its dependency paths, attention masks,
synthesis shifts, and multiscale delays, which do not depend on the learned
parameter values. Random
weights also reduce the possibility that an invalid dependency is hidden by a
degenerate zero mapping. In both experiments, the perturbed suffix produces
large changes in later logits, confirming that the perturbations are
effective.

\subsection{Causality Results}
\label{app:causality-results}

We perform two tests. The \emph{midpoint test} uses cutoff $t=1024$ and
changes every byte in positions 1025--2047. The \emph{exhaustive test} repeats
the experiment for every cutoff $t\in\{0,\ldots,2047\}$.

Table~\ref{tab:causality-results} reports the results. Repeated execution of
the identical input produces exactly the same logits in both tests:
$\delta_{\mathrm{rep}}=0$. In the midpoint test, the maximum shared-prefix
difference is also exactly zero, while the changed suffix produces a maximum
logit difference of $4.3369$.

Across all 2,048 exhaustive cutoffs, the maximum shared-prefix discrepancy is
$5.0664\times10^{-6}$, below the $10^{-5}$ acceptance threshold. The largest
difference occurs at cutoff $t=1362$, and no cutoff exceeds the threshold.
Meanwhile, the maximum changed-suffix difference is $5.4749$, showing that the
future perturbations materially change later predictions while leaving earlier
predictions invariant within numerical tolerance.

\begin{table}[t]
\centering
\caption{
Future-perturbation causality tests for the 2,048-byte all-scale fusion configuration
under controlled FP32 execution. Prefix differences must be at most
$10^{-5}$. Lower prefix difference is better.
}
\label{tab:causality-results}
\small
\setlength{\tabcolsep}{3.5pt}

\begin{tabular}{lrrrrc}
\toprule
\textbf{Test}
& \textbf{Cutoffs}
& $\boldsymbol{\delta_{\mathrm{rep}}}$
& \textbf{Max prefix diff.}
& \textbf{Max suffix diff.}
& \textbf{Result} \\
\midrule
Midpoint
& 1
& $0$
& $0$
& $4.3369$
& Pass \\
Exhaustive
& 2,048
& $0$
& $5.0664{\times}10^{-6}$
& $5.4749$
& Pass \\
\bottomrule
\end{tabular}

\end{table}

These results empirically validate the span-based argument for the tested
configuration, including the FastWeight analysis blocks, learned lossless
lifting steps, shifted super-resolution synthesis, semantic skip connections,
carrier features, all-scale semantic and detail fusion, and final causal
attention. The shared-state bottleneck in \textsc{DeepMem} satisfies the
same causal invariant by composition of causal FastWeight updates; the numeric
future-perturbation measurements reported here concern the all-scale fusion
comparison configuration.

\clearpage


\end{document}